\def\year{2019}\relax
\documentclass[letterpaper]{article} 
\usepackage{aaai19}  
\usepackage{times}  
\usepackage{helvet}  
\usepackage{courier}  
\usepackage{url}  
\usepackage{graphicx}  
\usepackage{color}
\usepackage{fancyhdr}

\newcommand{\ignore}[1]{}

\usepackage[a4paper,margin=1in]{geometry}

\usepackage{tabularx}
\newcolumntype{R}[1]{>{\hsize=#1\hsize\raggedleft\arraybackslash}X}%
\newcolumntype{L}[1]{>{\hsize=#1\hsize\raggedright\arraybackslash}X}%
\newcolumntype{C}[1]{>{\hsize=#1\hsize\centering\arraybackslash}X}%

\begin{document}
%
\title{Named Person Coreference in English News}
\author{ Oshin Agarwal\thanks{equal contribution}, Sanjay Subramanian\footnotemark[1], Ani Nenkova, Dan Roth \\
Department of Computer and Information Science, University of Pennsylvania
}
\maketitle
\begin{abstract}

People are often entities of interest in tasks such as search and information extraction. In these tasks, the goal is to find as much information as possible about people specified by their name. However in text, some of the references to people are by pronouns (she, his) or generic descriptions (the professor, the German chancellor). It is therefore important that coreference resolution systems are able to link these different types of mentions to the correct person name. Here, we evaluate two state of the art coreference resolution systems on the subtask of Named Person Coreference, in which we are interested in identifying a person mentioned by name, along with all other mentions of the person, by pronoun or generic noun phrase. Our analysis reveals that standard coreference metrics do not reflect adequately the requirements in this task: they do not penalize systems for not identifying any mentions by name and they reward systems even if systems find correctly mentions to the same entity but fail to link these to a proper name (she--the student---no name). 
We introduce new metrics for evaluating named person coreference that address these discrepancies. We present a simple rule-based named entity recognition driven system, which outperforms the current state-of-the-art systems on these task-specific metrics and performs on par with them on traditional coreference evaluations. Finally, we present similar evaluation for coreference resolution of other named entities and show that the rule-based approach is effective only for person named coreference, not other named entity types.\ignore{\footnote{Our system and the relevant OntoNotes data will be made publicly available upon publication of this paper.} }

\end{abstract}

\section{Introduction}
Coreference resolution is the task of identifying all expressions in text that refer to the same entity.  In this paper we set out to provide an in-depth analysis and improve performance on a special coreference subtask: finding all references---either by name, pronoun or nominal---to a person named in the text. Named Person Coreference (NPC) would be especially useful for downstream information extraction tasks, in which facts about the person are extracted from large textual corpora and used as knowledge source for a range of artificial intelligence systems.  NPC extends the power of Named Entity Recognition (NER) systems by not only finding all text snippets that are a person's name, but also identifying all places in the text where the same person was mentioned. 

Our work is oriented towards practical uses of the results in downstream applications, so we re-examine standard coreference metrics and find them lacking in their ability to quantify the performance of systems. Since applications require information about people and people are identified by their names, the evaluation metrics for this NPC task should focus on the resolution of mentions to the correct name. If all the pronouns referring to a person are resolved correctly to each other but are not linked to any named mention or are linked to a wrong named mention, their correct resolution to each other would not be useful for downstream applications. Standard coreference metrics do not incorporate these aspects of performance and hence give high performance for results unsuitable for further use. We also show that the existing metrics are not sensitive to finding any mention to a person at all. They give higher values for systems that do not find a large number of entities but do good coreference resolution on the subset of entities they find. 

We introduce new metrics to overcome these shortcomings. We separate the identification of entities and resolution of different mention types, thus transparently tracking areas of system performance and improvement. We create a subset of the standard coreference resolution annotated data sets by identifying all named person entities and adapting existing coreference systems to filter their original output to only person entities identified by name. 

Inspired by the error analysis of the performance of current coreference systems, we introduce a new solution to the NPC task that builds upon the capabilities of state of the art named entity recognition systems. We describe a highly effective rule-based approach to named person coreference that combines NER-driven rules and basic heuristics based on writing styles for resolving pronominal coreference. We show that this system has comparable performance to the state of the art systems on the standard metrics and superior performance on the task-informed metrics we introduce.

Finally, we test our approach for named entity coreference on entities other than people. We find that the state-of-the-art coreference systems do not suffer from the same issues for other entities as they do for persons and thus the results are similar across both standard metrics and NPC metrics. 

\section{Why Named Person Coreference?}

People, and the information about people expressed in text, are of interest in numerous applications. Moreover, references to people behave differently from references to other named entities, indicating that the intersection of coreference and mentions to people represents a fruitful area for application-oriented research. We expand on these observations motivating our work.  

\subsection{NPC in Downstream Applications}

Many information extraction and language technology tasks involve people. About 15\% of all web searches contain a person's name \cite{DBLP:conf/sigir/WeerkampBKMBR11} and especially in news search, NPC can help find articles in which the person of interest is the focus of discussion, mentioned by pronouns in addition to their full name.\footnote{An open problem in search is to disambiguate between different people with the same name, mentioned in different documents \cite{Artiles2010WePS3EC}. We do not deal with this problem and resolve mentions only within the same document.}

People are also often targets for information extraction systems \cite{Ji:2011:KBP:2002472.2002618} and for knowledge base completion tasks \cite{west2014knowledge}. Yet about half of the references to a person in text are not by name (cf. the first column of Table \ref{StatsOtherEntities}). Systems need to extract information about the person where they are mentioned explicitly by name as well as referenced by a pronoun. 

Biography summarization \cite{zhou2005multi} needs to extract sentences from the news that contain information about a given person. More relevant sentences can be extracted if we know which pronouns and nominals refer to this person. Similarly, creation of proper noun ontologies \cite{mann2002fine} can use patterns other than (proper noun - common noun) if other references to the entity are known.

\ignore{There has also been interest in coreference resolution for person entities specifically in the context of video transcripts \cite{ramanathan2014linking}. }

\begin{table}[t]
\centering
\small
\setlength{\tabcolsep}{4pt}
\begin{tabularx}{\linewidth}{L{1.5}C{0.4}C{0.4}C{0.4}C{0.4}}
\hline
&{\bf PER}&{\bf ORG}&{\bf GPE}&{\bf DATE}\\\hline 
Non-singleton &67.98\%&50.76\%&52.02\%&21.3\%\\
Entities &15.44\%&11.35\%&11.36\%&4.24\%\\
Entity Mentions&24.47\%&14.33\%&13.62\%&3.41\%\\
Named mentions&45.12\%&55.79\%&68.43\%&82.69\%\\
Non-named mentions&54.88\%&44.21\%&31.57\%&17.31\%\\
Avg cluster size&5.74&4.57&4.34&2.89\\\hline
\end{tabularx} 
\caption{Statistics, in OntoNotes (nw,bn,mz), on coreference for PERson, ORGanization, geopolitical entity (GPE) and DATE named entities.  Non-singleton entities are mentioned at least twice in a text and so require coreference. Entities is the percentage of coreference clusters to entities of the given type. Entity mentions is the percentage of all individual references to any entity of a given type. Average cluster size is the average number of coreferent mentions.}\label{StatsOtherEntities}
\end{table}

\subsection{Coreference and Named Entities}
References to people are distributed quite differently from other types of named entities, as we show in Table \ref{StatsOtherEntities}. To compile these numbers, we make use of the OntoNotes coreference resolution corpus \cite{pradhan2007ontonotes} and gold-standard annotations for named entity recognition on the same data. In this way, we can quantify the patterns in coreference of different named entity types. 


People are usually identified by their name in text:  86\% of the coreference clusters on people in the OntoNotes training data have at least one named mention; 88.2\% of animate third person singular pronouns (pronouns that can be used to refer to people) are a part of coreference chains with a named person, as seen in Table ~\ref{PronounStats}.\footnote{Female pronouns occur considerably less often than male pronouns, and a larger portion of female pronouns are to unnamed women. These numbers raise question about usage but overall the trend is clear: third person animate pronouns, in their vast majority, resolve to a person named in the article.} Therefore, it is uncommon to refer to a person without identifying them by their name at least once. 

All named entities are on average much less likely to be singletons than a typical entity, mentioned only once in the text and not requiring coreference resolution \cite{DeMarneffe:2015:MLD:2831407.2831417}, and of the named entities people are most likely to be mentioned repeatedly. People are rarely mentioned only once in news (see the first line in Table \ref{StatsOtherEntities}): 68\% of people named in text have at least one other coreferent mention to them in contrast to 51\% and 52\% for organizations and locations respectively. 

Named people entities make up 15\% of all coreference clusters in OntoNotes (see line 2 of Table \ref{StatsOtherEntities}), yet 25\% of all mentions that require coreference resolution are mentions of people (Line 3 in Table \ref{StatsOtherEntities}). There are more mentions of the same person on average (line 6 in Table \ref{StatsOtherEntities}) and PERson is the entity type with the largest portion of references that are not by name (line 5 in Table \ref{StatsOtherEntities}). Only 45\% of mentions to people are by name, compared to 56\% for organizations and 68\% for geo-political entities and over 80\% for dates. 

In sum, references to named people make up a quarter of all references involved in coreference resolution and only half of these references are by name, so output of traditional NER systems that identify the names are not sufficient to track all mentions of the person; rules to disambiguate pronouns and noun phrase references are needed to track all mentions of people. 

\begin{table}[t]
\centering
\small
\setlength{\tabcolsep}{4pt}
\begin{tabularx}{0.7\linewidth}{C{0.25}|C{0.25}|C{0.4}}
&{\bf Total}&{\bf Part of PER cluster}\\\hline
he&3062&2705 (88.3\%)\\
him&447&379 (84.7\%)\\
his&1931&1669 (86.4\%)\\\hline
she&718&588 (81.8\%)\\
her&667&521 (78.1\%)\\
hers&2&2 (100\%)\\\hline
\end{tabularx} 
\caption{Third person singular pronouns in the Ontonotes train set.}\label{PronounStats}
\end{table}

\section{NPC and Coreference Evaluation}
\label{sec:generalmetrics}

Named Person Coreference is motivated by the needs in downstream applications. This setting allows us to critically review current practices for coreference evaluation and identify aspects where they fall short in quantifying system performance. 

Shared tasks on coreference (at CoNLL-2011 and 2012 \cite{pradhan-EtAl:2014:P14-2} ) use the average of three scores as their official evaluation: MUC \cite{Vilain:1995:MCS:1072399.1072405}, \(B^3\) \cite{DBLP:conf/acl/BaggaB98} and CEAFE \cite{DBLP:conf/naacl/Luo05}.  \ignore{The first two scores,  MUC \cite{Vilain:1995:MCS:1072399.1072405} and \(B^3\) \cite{DBLP:conf/acl/BaggaB98}, assess system performance in terms of their ability to decide if a pair of mentions (noun phrases) in text refer to the same entity and  mention-focused recall/precision respectively. Below we describe these in more detail and point out how they are deficient when examined in the context of downstream tasks, particularly ones that involve NPC. The third score, CEAFE \cite{DBLP:conf/naacl/Luo05}, instead incorporates the notion of an entity as the key element in the evaluation.} Prior work \cite{moosavi2016coreference} discussed the shortcoming of these metrics and introduced the link entity aware (LEA) score to tackle them. Below we describe these and point out how they are deficient when examined in the context of downstream tasks, particularly ones that involve NPC. \footnote{This view of evaluation focused on the ability of a system to discover and track entities, which we later expand on, is similar to the vastly popular entity linking task, in which named mentions in text are linked to an abstract entity, such as one defined in Wikipedia \cite{Mihalcea:2007:WLD:1321440.1321475,DBLP:conf/acl/HanS11,DBLP:journals/tacl/DurrettK14,DBLP:conf/naacl/PanCHJK15,Radhakrishnan2018ELDENIE}.}

\begin{table}[t]
\centering
\small
\setlength{\tabcolsep}{4pt}
\begin{tabularx}{\linewidth}{L{}}
\\\hline
Gold clusters: \(\{John Doe, he_5, he_6, he_7, he_8, he_9\},\)\\
\(\{Richard Roe, he_1, he_2\}, \{Joe Smith, he_3, he_4\},\)\\\hline
Solution 1: \(\{John Doe, he_5, he_6, he_7, he_8, he_9\}\)\\
Solution 2: \(\{he_1, he_2\}, \{he_3, he_4\}, \{he_5, he_6, he_7, he_8, he_9\}\) 
Solution 3: \(\{Richard Roe, he_1\}, \{Joe Smith, he_3\},\)\\
\(\{John Doe, he_5, he_6\}\)\\\hline
\end{tabularx} 
\caption{NPC Examples}\label{Examples}
\end{table}


\paragraph{MUC}
MUC computes performance at the entity level, where a goldstandard cluster of mentions (noun phrases in text) represents an entity. The recall for an entity is based on 
the minimum number of links that would have to be added in the predicted clusters containing any mention of this entity, to make them connected and part of the same cluster. Precision is computed by reversing the role of gold and predicted clusters. 

\paragraph{B-cubed}
\(B^3\) measures performance on the mention level. It iterates over all goldstandard mentions of an entity, averaging the recall of its gold cluster in its predicted cluster. Different mentions may have been predicted as referring to different entities, placed in different solution clusters. It computes precision by reversing the role of gold and predicted clusters. 


\paragraph{CEAF}
CEAF first finds a one-to-one mapping between gold and predicted clusters, in effect placing the discovery of entities first. It then computes recall as the number of same or similar mentions shared by the gold and predicted clusters divided by the number of mentions in the gold cluster. Precision similarly is equal to the number of same/similar entities in the gold and predicted, divided by the number of mentions in the predicted cluster. The final version averages these number either per mention as in B-cubed (CEAFm), or per entity (CEAFe), as would be more reasonable. 

\paragraph{LEA}
LEA is a link-based metric most similar to MUC, that computes recall as the number of correctly resolved links between mentions, weighting the results for each entity by its number of mentions. In this way, resolving correctly an entity with more mentions contributes more to the overall score than resolving correctly all mentions to an entity mentions only twice for example. Precision is computed by reversing the role of gold and predicted clusters. 

\begin{table}[t]
\centering
\small
\setlength{\tabcolsep}{2pt}
\begin{tabularx}{\linewidth}{C{0.35}|C{0.13}C{0.13}C{0.13}|C{0.13}C{0.13}C{0.13}|C{0.13}C{0.13}C{0.13}}
&\multicolumn{3}{>{\hsize=\dimexpr3\hsize+4\tabcolsep+1\arrayrulewidth\relax}C{0.13}|}{\bf Solution 1}&\multicolumn{3}{>{\hsize=\dimexpr3\hsize+4\tabcolsep+1\arrayrulewidth\relax}C{0.13}|}{\bf Solution 2}&\multicolumn{3}{>{\hsize=\dimexpr3\hsize+4\tabcolsep+1\arrayrulewidth\relax}C{0.13}}{\bf Solution 3}\\\cline{2-10}
&{\bf R}&{\bf P}&{\bf F1}&{\bf R}&{\bf P}&{\bf F1}&{\bf R}&{\bf P}&{\bf F1}\\\hline
MUC&0.55&1&0.71&0.66&1&0.8&0.44&1&0.61\\
B-cub&0.5&1&0.66&0.56&1&0.72&0.34&1&0.51\\
CEAFm&0.5&1&0.66&0.75&1&0.85&0.58&1&0.73\\
CEAFe&0.33&1&0.5&0.83&0.83&0.83&0.75&0.75&0.75\\
LEA&0.5&1&0.66&0.5&1&0.66&0.26&1&0.42
\end{tabularx} 
\caption{Evaluation of the hypothetical solutions on NPC examples in Table \ref{Examples}}\label{ExampleStats}
\end{table}

Recall that the goal of NPC is to find all mentions referring to a person, identified by their name. We provide a made up example of a goldstandard and three possible solutions in Table ~\ref{Examples}. The gold standard contains three entities: John Doe, Richard Roe and Joe Smith. As the clusters indicate, each is also mentioned by a pronoun a number of times. The first hypothetical system response (solution 1) identifies only one entity.\ignore{, decides that there is only one entity mentioned in the text.} It finds all mentions to John Doe correctly but completely misses all mentions to the other two entities. The second solution correctly resolves the pronouns to each other but does not link these to any names. The third solution is able to identify a few mentions for all of the three entities. 
Intuitively, solution 2 has little practical value, solution 3 is best and solution 1 is acceptable.

The results for standard coreference scores are shown in Table ~\ref{ExampleStats}.  All of the metrics have the highest values for the Solution 2, which does not identify a single name. The same would be true even if any of these correct set of pronouns were linked to the wrong name. This is because none of the metrics take into account the types of mentions and the need for resolution to correct names. 


\section{NPC evaluation metrics}
\label{sec:npcmetrics}
We showed in the previous section that the existing coreference metrics are not suitable for the NPC task. Moreover, having a single number to track all aspects of the systems, makes it less interpretable. Prior work has argued that when coreference is used in downstream applications, evaluation criteria should be interpretable \cite{tuggener2014coreference}. We concur, and introduce a set of task-specific criteria for evaluation of NPC. These are inspired by error analysis we performed. For clarity, we provide examples errors on Ontonotes.

\subsection{Entity F1: Matching Output to Goldstandard Entities}

In the goldstandard, all noun phrases referring to the same person mentioned by name at least once are grouped into entities. In the system output, we also wish to find the chain corresponding to each person, similar to the motivation of the CEAFE evaluation. To map entity chains between the goldstandard and the system output, we select for each goldstandard entity, the predicted entity that has the highest F1 score with respect to the mentions it contains. The entity F1 score is the harmonic mean of the precision\footnote{Number of mentions to the goldstandard entity also in the system entity divided by the number of all system mentions.} and recall\footnote{Number of mentions to the goldstandard entity also in the system entity divided by the number of mentions in the goldstandard entity.} for the entity mapping. 

To compute the intersection between a goldstandard and a system entity, we first augment each goldstandard entity with a list of all variations of the person's name. We rely on the goldstandard named entity annotation in OntoNotes and intersect this with the membership in a coreference chain. This provides lists of the full name, last name, occasionally nicknames and variants of the name, i.e. {\em \{Frank Curzio, Francis X. Curzio, Curzio\}}, {\em \{Dwayne Dog Chapman, Dog Chapman, Chapman\}}. We consider a predicted chain to be a candidate match for a gold chain only if it contains at least one of the name variants. 

We consider a predicted mention to match if it matches the gold mention exactly in the same place of the text. Mention detection has a huge impact on evaluation but we use an exact span matching to be consistent with the existing coreference scorers.\footnote{Exact mention match is used for calculating F1. We do not use exact mention match to find candidate chains as the presence of the name can indicate which person the cluster is about.}

If a goldstandard entity does not get paired with any system entity, the F1 for that entity is taken to be zero. In our example above, Solution 1 will have poor recall, because it finds only one entity in a document containing three entities. We find the overall F1 of the system as the average of the F1 for each gold entity.

\subsection{Entity not found}

The entity F1 evaluation gives a sense of overall system performance but mixes true purity of the system-discovered entities and the ability to discover entities at all.
"Entity not found" is the error when no NPC system output overlaps with a gold standard entity. Entities not found contribute a score of 0 for the average F1.\footnote{We consider only chains containing a named mention. Chains that do not contain any named mention are filtered out. More details on filtering to follow in the section on system performance.} 

\subsection{Pronoun Resolution Accuracy}
Finally, we track the entity F1 when only mentions of given syntactic type are preserved in the chain---name, pronoun and nominal. Of special interest is to track system performance when resolving pronouns. Many of these issues arise due to the need for commonsense knowledge and reasoning for correct resolution, as in the these examples:

\noindent
\textit{\(\textbf{A school official}_1\) talked about the Vice President's chances during an interview with the Boston Globe. \(\textbf{He}_2\) says it's unlikely Gore will be selected, because \(\textbf{he}_3\) doesn't have enough experience in the academic world.}

The pronoun \(\textit{he}_3\) is incorrectly resolved to the school official instead of Al Gore. To correctly resolve this, we need to know that not having enough experience would be the reason for Gore not getting selected and not the reason for the school official making a statement.

\noindent
\textit{Maybe Lily became so obsessed with where people slept and how because her own arrangements kept shifting. When \(\textbf{Rosie}_1\) died, \(\textbf{her}_2\) uncles moved in and let \(\textbf{her}_3\) make the sleeping and other household arrangements.}

Rosie is Lily's mother (explained earlier in the article). Both \(\textit{her}_{2,3}\) are incorrectly resolved to Rosie, despite the common sense fact that a person can not make household arrangements after their death.

\begin{table*}[t]
\centering
\small
\setlength{\tabcolsep}{4pt}
\begin{tabularx}{\linewidth}{L{1.2}C{0.31}C{0.31}C{0.31}C{0.31}|C{0.31}C{0.35}C{0.35}|C{0.28}C{0.28}C{0.28}}
\hline
&{\bf Chains not found}&{\bf F1 (NPC)}&{\bf Avg F1 (coref)}&{\bf LEA F1}&{\bf F1 (names)}&{\bf F1 (pronouns)}&{\bf F1 (nominals)}&{\bf F1 (nw)}&{\bf F1 (bn)}&{\bf F1 (mz)}\\\hline
CoreNlp deterministic&{\bf 24.76\%}&0.501&0.499&0.409&0.489&0.384&0.027&0.499&0.494&0.525\\
CoreNlp statistical&39.30\%&0.45&0.577&0.499&0.488&0.307&0.019&0.411&0.49&0.45\\
CoreNlp neural&27.35\%&{\bf 0.572}&0.677&0.613&{\bf 0.612}&{\bf 0.397}&0.062&{\bf 0.59}&0.58&0.502\\
AllenNlp&31.87\%&0.563&{\bf 0.71}&{\bf 0.66}&0.61&0.349&{\bf 0.078}&0.412&{\bf 0.699}&{\bf 0.608}\\\hline
\end{tabularx} 
\caption{Performance of existing systems. The left panel shows named people coreference metrics (percentage chains not found and entity F1), and average F1 coreference evaluation combining MUC, \(B^3\) and CEAFE, on all test data. The middle panel shows entity F1 by type of mention, names, pronouns or nominals. The right panel shows entity F1 broken down by subgenres in the test data: newswire (nw), broadcast news (bn) and magazines (mz). Existing F1 metric is not sensitive to chains not found, and this explains the difference from the new intuitive NPC metric. For each metric, the best system has been bold-faced}\label{AllPeopleMetrics}
\end{table*}

\subsection{Over-splitting and over-combination of entities}
Sometimes, systems produce more than one clusters, each containing the same name. In this case we say that the gold-standard entity was over-split. Following is an example illustrating this error - 

\noindent
\textit{\(\textbf{Frank Curzio}_1\). Many people now claim to have predicted the 1987 crash. \(\textbf{Queens newsletter writer Francis X. Curzio}_2\) actually did it: \(\textbf{He}_3\) stated in writing in September 1987 that the Dow Jones Industrial Average was likely to decline about 500 points the following month. \(\textbf{Mr. Curzio}_4\) says what happens now will depend a good deal on the Federal Reserve Board. If it promptly cuts the discount rate it charges on loans to banks, \(\textbf{he}_5\) says, '' That could quiet things down. ''}

The above text mentions Frank Curzio 5 times, however all the mentions aren't considered to be coreferent to each other. Mentions 2 and 3 form one cluster while the other three form another one.

At times systems also produce coreference chains that combines mentions to two different people. This error occurs when different people mentioned in the text have the same last name (notoriously for U.S. news Bill Clinton and Hilary Clinton) but also occasionally in cases where the names are completely different but the roles of the people are similar, as in the example below - 

\noindent
\textit{\textbf{UN Secretary General Kofi \(\textbf {Annan}_1\)} said Wednesday, it is important to help spread democracy around the world.
VOA's Breck Ardery reports from the United Nations . In a new report, \(\textbf{the Secretary General}_2\) says democratization has now taken root as a universal norm and that the United Nations should strengthen its commitment to assisting nations that are moving toward democracy. Commenting on the report, \textbf{UN Assistant Secretary General for Political Affairs Danilo \(\textbf {Turk}_3\)} told reporters the principle of national sovereignty does not preclude support for democracy.}

Although it is clear from the above text snippet that Kofi Annan and Danilo Turk are two different people, they have similar titles and recognized to be coreferent.

Following is another document, where Scott Peterson and his wife Laci Peterson are said to be coreferrent, in spite of having distinct gender pronouns referring to each. \ignore{Apart from the highlighted mention, Scott Peterson in the second to last sentence isn't detected, and 'Peterson' and 'his wife Laci' in the last sentence form singleton clusters each.}

\noindent
\textit{Moving on to the \(\textbf{Laci Peterson}_1\) case Bill, an unusual request from \(\textbf{Scott Peterson}_2\)'s attorney. A pair of \textbf{Laci \(\textbf {Peterson}_3\)}'s missing shoes could be very important evidence in \(\textbf{Scott's}_4\) murder trial. \(\textbf{She}_5\) is asking anyone who finds the pair to give them back. No sign of Geragos or \(\textbf{Peterson}_6\) in court yesterday when a judge was considering whether to unseal warrants obtained before Scott Peterson's arrest. Peterson awaiting trial in the murder of \(\textbf{his}_7\) wife Laci and their unborn son.}

\section{Evaluation of existing systems}
\label{sec:evals}

We evaluate the Stanford coreference system, with its deterministic \cite{DBLP:conf/emnlp/RaghunathanLRCSJM10}, statistical \cite{DBLP:conf/acl/ClarkM15} and neural \cite{DBLP:conf/emnlp/ClarkM16} versions, and the neural end-to-end AllenNLP system \cite{DBLP:conf/emnlp/LeeHLZ17} on both the existing metrics and the NPC metrics.

These general coreference systems find coreferring expressions of any type and produce coreference clusters for all mentioned entities (groups  of noun phrases that refer to the same entity). In named person coreference, the goal is to find all mentions to {\em a person} who has been {\em referred to by name} at least once in the document. This means that the output of off-the-shelf coreference systems has to be filtered to keep only chains that contain at least one mention noun phrase with a syntactic head that is a person's name. For our evaluation, we use dependency parsing to detect whether a name is the head of a mention, by checking that no other word in the mention is an ancestor of the name in the dependency parse tree. We use automatic {\sc person} NER tags from Stanford coreNLP  \cite{DBLP:conf/acl/FinkelGM05} to determine if the head is a name. We call the coreference chains remaining after filtering {\em NPC entities}.

Less strict filtering, such as the presence of 3rd person singular personal or possessive pronouns would also indicate that the corresponding entity is a person. For NPC, we insist on having at least one named mention,\ignore{to facilitate search and information extraction,} as elaborated upon in the earlier sections.  We found that the AllenNLP system does not have a named mention in about 30\% of the coreference chains that do contain a personal or possessive third person pronoun. This number is about 20\% for the CoreNLP neural system. 

For evaluating the systems, we use only the relevant subsample of the standard coreference evaluation data in OntoNotes. We work with the test newswire (89 documents, 136 NPC chains), broadcast news (93 documents, 135 NPC chains) and magazine (45 documents, 49 NPC chains) documents only. We evaluate on the 320 NPC chains in these 227 documents. The NPC chains contain three types of mentions---names, pronouns, and nominals. Nominals account for less than 5\% of the mentions in all genres, while the remaining mentions are split almost equally between  names and pronouns.  

The third column in the first panel of Table~\ref{AllPeopleMetrics} shows the standard F1 on all the systems.  As expected, the AllenNLP systems outperforms all the systems, with CoreNLP neural as a close second. Both perform much than the CoreNLP deterministic and statistical systems.

However, this difference in performance isn't as big on the NPC F1 (see second column of Table \ref{AllPeopleMetrics}). The Stanford CoreNLP neural system has the highest NPC F1, only slightly better than the AllenNLP system.\footnote{This evaluation uses an exact span matching to be consistent with the existing coreference scorers. Our experiments showed a relaxed mention span matching allowing a difference of a few words results in gains up to 12 points F1. The gain with relaxed mention matching is highest for the deterministic system which ends up performing as good as the CoreNLP neural system.} The NPC F1 includes 0s for goldstandard entities not found by the system. We track this separately as well (first column of Table \ref{AllPeopleMetrics}). The deterministic and neural Stanford CoreNLP systems have the lowest percentage of entities not found errors, about a quarter of all entities. The Stanford statistical system is worst at finding entities, missing almost 40\% of entities.

We also separate the performance of the systems by mention type. The second panel of Table~\ref{AllPeopleMetrics} reveals that the systems make more mistakes on pronouns compared to names. 

We also evaluated separately on the three genres, as shown in the third panel of Table \ref{AllPeopleMetrics}. While the CoreNLP system outperforms on newswire by a huge margin, the AllenNLP system performs the best on broadcast news and magazines. Like overall F1, scores are greatly impacted by the mention detection and the difference in performance becomes less with a relaxed mention matching.

We tracked the over-splitting and the over-combination of entities as well. However, the overall values were quite small and similar for all the systems and have thus not been included in the results here.

\begin{table*}[t]
\centering
\small
\setlength{\tabcolsep}{4pt}
\begin{tabularx}{\linewidth}{L{1.2}C{0.31}C{0.31}C{0.31}C{0.31}|C{0.31}C{0.35}C{0.35}|C{0.28}C{0.28}C{0.28}}
\hline
&{\bf Chains not found}&{\bf F1 (NPC)}&{\bf Avg F1 (coref)}&{\bf LEA F1}&{\bf F1 (names)}&{\bf F1 (pronouns)}&{\bf F1 (nominals)}&{\bf F1 (nw)}&{\bf F1 (bn)}&{\bf F1 (mz)}\\\hline
NER-DE&{\bf 12.50\%}&{\bf 0.685}&0.67&0.583&{\bf 0.727}&0.477&0.012&{\bf 0.601}&{\bf 0.739}&{\bf 0.771}\\
NER-DE (CoreNlp mentions)&13.12\%&0.594&0.584&0.481&0.58&0.466&0.028&0.516&0.663&0.619\\
NER-DE (CogComp mentions)&13.12\%&0.616&0.594&0.484&0.598&{\bf 0.479}&0.006&0.516&0.702&0.656\\
NER-DE (AllenNlp mentions)&35.31\%&0.541&{\bf 0.676}&{\bf 0.606}&0.588&0.335&{\bf 0.032}&0.396&0.675&0.576\\
NER-DE (gold mentions)&11.56\%&0.789&0.817&0.756&0.867&0.488&0.056&0.749&0.815&0.831\\\hline
\end{tabularx} 
\caption{Performance of NER-DE system with different methods of mention detection. The left panel shows named people coreference metrics (percentage chains not found and entity F1), and average F1 coreference evaluation combining MUC, \(B^3\) and CEAFE, on all test data. The middle panel shows entity F1 by type of mention, names, pronouns or nominals. The right panel shows entity F1 broken dawn by subgenres in the test data: newswire (nw), broadcast news (bn) and magazines (mz). Existing F1 metric is not sensitive to chains not found, and this explains the difference from the new intuitive NPC metric. For each metric, the best system has been bold-faced}\label{NERDEPeopleMetrics}
\end{table*}

\section{NER-driven Coreference}

The most striking error of current systems is their inability to find a substantial fraction (20\%--40\%) of the NPC entities. In these cases the system may have produced an entity that overlaps with some mentions but does not include the name. Detection of named mentions can be done with high accuracy by named entity recognition systems \cite{DBLP:conf/acl/StoyanovGCR09} and the matching of names can also be done accurately via string matching \cite{Wacholder:1997:DPN:974557.974587,wick2009entity}. Pronoun resolution is more difficult and sometimes even requires background knowledge. However, generally, writing styles dictate how pronouns are used in text. Nominal references, on the other hand, are unlikely to make any noticeable difference since there are only a handful. Inspired by these observations, we present a NER-driven rule based coreference system, NER-DE.  We perform clustering on named entities and link pronouns using basic heuristics. This is simpler than prior work on coreference resolution using clustering. Authors of \cite{cardie1999noun} represent all noun phrases using a set of features and perform clustering on these representations, which can also lead to clusters not having an associated name.

\paragraph{People-mention coreference}
In a first pass, NER-DE finds all spans of text that are {\sc person} named entities, using the Cogcomp NER \cite{RatinovRo09}. We later use a dependency parser \cite{honnibal-johnson:2015:EMNLP} to find the noun phrase of which that named entity serves as the syntactic head. This is the smallest span that contains all descendants of the last word in the named entity. We also use two additional rules: (1) if there is an 'and' immediately after the name, the end of the name is the end of the mention, and (2) if the token immediately before the name is a noun or a proper noun, we include that token and its descendants (in the parse tree) in the mention span. In addition, we experimented with taking mention detection from existing coreference systems rather than using this dependency parsing-based method.

Coreference decisions are done on the named entity spans. We start by focusing on names, initializing a separate chain for each named entity. We then do agglomerative clustering of the names, using a named entity similarity metric~\cite{cogcompnlp2018lrec}. We merge two chains if the longest mention of the first chain has a similarity of more than 0.5 with any mention of the second chain. Most similarity scores are close to 0 and 1 so the system is not sensitive to the chosen similarity cut off.

\paragraph{Resolving pronouns}
We add first-person and third-person singular pronouns to the coreference chains containing named mentions using three rules, restricted by gender compatibility. We consider the gender of the chain to be the gender of the first word in the longest named mention of the chain. We determine the gender using a gazetteer of names, created in an unsupervised manner from Wikipedia. For each person's page, the name is considered female if the number of female pronouns is greater than the number of male pronouns in the first paragraph of the page, else it is considered male. A name is considered unisex if it does not appear in the gazetteer. This gazetteer has been evaluated from names that originate from different languages and gives accuracies in the high 90s. We use the following rules for pronoun resolution:

{\em (i)} If the pronoun is the subject of a verb and the preceding subject is a name, we assign the pronoun to that name. For instance, \textit{he} is assigned to \textit{Henry} in the text - \textit{Henry went to see Barry in the hospital. Afterward, he ate a pizza.}, 

{\em (ii)} Else, we assign the pronoun to the nearest preceding name such that {\em (a)} the pronoun is not a subject of a verb whose object is the name (or any other name in the same entity cluster) {\em (b)} the pronoun is not an object of a verb whose subject is the name (or any other name in the same chain), and {\em (c)} the name appears at least once in the 100 words preceding the pronoun.

{\em (iii)} If no name satisfies any of the above two conditions, the pronoun is not assigned to a name. 

We remove chains that have only one mention, to follow the convention of having only non-singleton chains.

\section{Evaluation of NER-DE}
We evaluate NER-DE on both standard and NPC metrics. The results can be seen in Table ~\ref{NERDEPeopleMetrics}. 

NER-DE performs better than the CoreNLP deterministic and statistical systems on the standard metrics. AllenNLP still outperforms all others on this metric. However, as discussed in the earlier sections, this metric is not sensitive to entities not found at all. NER-DE misses 12.5\% of the chains, which is much smaller than what the other systems miss and is still comparatively weaker on the standard metrics. 4.67\% of the chains are missed because NER is not able to find a single named mention in that chain. The rest of the chains are missed due to the coreference algorithm, with no other reference linked to the name.

The existing metrics also do not take into account the resolution of mentions to the correct name, which is incorporated in the NPC F1. NER-DE outperforms all systems on NPC F1, both overall and in the subgenres of nw, bn and mz. It also does better on finding named mentions as well as pronouns. We should note that this improved performance is obtained by basic heuristics of matching pronouns to the nearest preceding named mentions with just a few additional constraints.

We also see that the performance varies by a significant amount depending on the source of mentions used by NER-DE. Surprisingly, the performance of NER-DE is better by using mentions generated using parsing, instead of using mentions generated by CogComp and CoreNLP systems, on both the NPC and standard metrics. The performance of NER-DE using AllenNLP mentions is comparable to parsed mentions using existing metrics but the AllenNLP systems combines mention detection with coreference and thus has better mentions. The NER-DE using parsed mentions still outperforms even the one using AllenNLP mentions on NPC metrics. Using gold mentions gives a very high performance on both the NPC and standard metrics, thus showing that NER-DE, which uses just NER and basic heuristics for pronominal resolution is surprisingly good at named person coreference. Improving mention detection will lead to even better performance.

\begin{table}[b]
\centering
\small
\setlength{\tabcolsep}{4pt}
\begin{tabularx}{0.85\linewidth}{L{0.6}C{0.38}C{0.35}C{0.35}}
\hline
&{\bf Chains not found}&{\bf F1 (NPC)}&{\bf Avg F1 (coref)}\\\hline
PER Allen&31.87\%&0.563&0.71\\
PER NER-DE&12.5\%&0.685&0.67\\\hline
ORG Allen&28.9\%&0.55&0.6\\
ORG NER-DE&28\%&0.41&0.42\\\hline
GPE Allen&13.9\%&0.75&0.75\\ 
GPE NER-DE&13.2\%&0.63&0.51\\\hline
\end{tabularx} 
\caption{Performance in coreference for PER, ORG, and GPE entities}\label{EvalOtherEntities}
\end{table}

\section{Coreference for Other NE types}
Natural language processing applications usually deal with PER, ORG and GPE entities together. Therefore, similar to PER, we evaluated the ORG and GPE coreference chains using standard and NPC metrics. We also built a similar NER-driven rule-based system with agglomerative clustering for resolving names and basic heuristics for resolving pronouns. \ignore{For the ORG and GPE entities, we did not use the dependency parsing-based rules for resolving pronouns but did use dependency parsing for selecting mention spans for named entities.} Unlike PER, the state-of-the-art coreference systems are able to find the same percentage of chains as the NER-driven system and do not suffer from the same issues as person entities that often. They are also able to resolve all types of mentions better than the NER-driven system. These results can be seen in Table \ref{EvalOtherEntities}. \ignore{The percentage of chains not found for the baseline system is a result of (1) NER recall and (2) singleton names, which are removed by the system. In order to determine the effect of each of this, we also computed the percentage of chains in which no entity was found by NER. These percentages are shown in Table \ref{EvalOtherEntities} for each entity. Singletons have a larger effect for each entity type, but the gap in percentage of chains not found between Allen and NER is largest for PER, followed by ORG and GPE.}

\section{Conclusion}
We presented the task of Named Person Coreference (NPC) and showed that the standard coreference metrics are not suitable for the evaluation of this task. We introduced interpretable evaluation metrics that tackle the shortcomings of the standard metrics and also track the different errors made by systems. We showed that the top off-the-shelf systems do not perform well on these metrics. They output many clusters without a link to any name or a link to the incorrect name. We showed that similar issues are not prevalent for other entities. We presented a simple NER-driven algorithm (NER-DE) for the named person coreference task that performs better than top off-the-shelf systems on the new metrics and on par on the existing metrics.

\bibliography{aaai}
\bibliographystyle{aaai}

\end{document}